# Facial Expression Recognition at the Edge: CPU vs GPU vs VPU vs TPU

Mohammadreza Mohammadi
mohammm@email.sc.edu
University of South Carolina
Columbia, SC, US

Heath Smith
smithmh6@email.sc.edu
University of South Carolina
Columbia, SC, US

Lareb Khan
lskhan@email.sc.edu
University of South Carolina
Columbia, SC, US

Ramtin Zand
ramtin@cse.sc.edu
University of South Carolina
Columbia, SC, US

## ABSTRACT

Facial Expression Recognition (FER) plays an important role in human-computer interactions and is used in a wide range of applications. Convolutional Neural Networks (CNN) have shown promise in their ability to classify human facial expressions, however, large CNNs are not well-suited to be implemented on resource- and energy-constrained IoT devices. In this work, we present a hierarchical framework for developing and optimizing hardware-aware CNNs tuned for deployment at the edge. We perform a comprehensive analysis across various edge AI accelerators including NVIDIA Jetson Nano, Intel Neural Compute Stick, and Coral TPU. Using the proposed strategy, we achieved a peak accuracy of 99.49% when testing on the CK+ facial expression recognition dataset. Additionally, we achieved a minimum inference latency of 0.39 milliseconds and a minimum power consumption of 0.52 Watts.

## 1 INTRODUCTION

Facial expressions are a vital part of human communication and make it possible to convey a wide range of complex emotions such as anger, happiness, sadness, or contempt. Humans are well-equipped with evolutionary tools that allow us to decode the facial expressions of other humans instinctively, but this is not such an easy task for computer vision systems [1]. Facial expressions can vary widely from one person to another and there is a great overlap between different classes of facial expressions. Using computer vision and machine learning (ML) systems to accomplish facial expression classification is beneficial to many applications which require complex human-machine interactions such as social robotics [2].

While ML systems can be trained to classify human facial expressions, this ability often comes at a large computational cost and poses challenges for edge computing systems and low-cost IoT devices [3]. Research in deep learning for computer vision applications has shown that increasing the depth of a convolutional neural network (CNN) model can increase the learning ability of the model. This is evident in very deep convolutional models such as ResNet [4], which is among the most popular models found in machine vision applications. The main drawback of these very deep models is the massive computational overhead required for a network containing tens of millions of parameters. Models such as *MobileNet*[5], attempt to reduce the large computational cost incurred in very deep networks by utilizing depthwise separable convolutional layers. However, it still can be difficult to efficiently deploy models of this size on low-cost edge devices with limited memory and power resources.

Developing efficient CNN models for resource- and energy-constrained IoT and edge devices can be challenging as network architectures are becoming larger and more complicated. Design spaces can include more than $10^{18}$ potential designs [6], leaving the human design process time-consuming and inefficient. Unlike manual design, automated ML (AutoML) [7] has been developed to automatically design neural networks with minimal human participation. AutoML effectively cuts down the skill barrier and broadens the application of neural networks by outperforming manually designed architectures on various tasks such as image classification [8], object detection [9], and semantic segmentation [10]. Neural architecture search (NAS) [11] and hyperparameter search are two categories of AutoML techniques for developing ML models. NAS approaches look across model descriptions of neural network designs, whereas hyperparameter search techniques attempt to identify the optimal values for hyperparameters given a fixed neural architecture. Hyperparameter search approaches use a variety of algorithms, including random search, Bayesian optimization [12], bandit-based methods [13], metaheuristics[14], and population-based training methods [15].

This work proposes an approach that aims to mitigate the large computational overhead of using deep CNNs for facial expression recognition by creating a hierarchical methodology that reduces the computational costs of accomplishing the task at edge devices while maintaining acceptable accuracy. In doing so, an improvement in hardware performance is expected when such optimized models are deployed on mobile devices. In this paper, we particularly optimize and deploy the facial recognition models on a tiny computer, i.e. Raspberry Pi, and three types of edge AI accelerators, i.e., graphic processing unit (GPU), tensor processing unit (TPU), and vision processing unit (VPU). A comprehensive comparison between different devices is provided in terms of accuracy, latency, and power consumption.

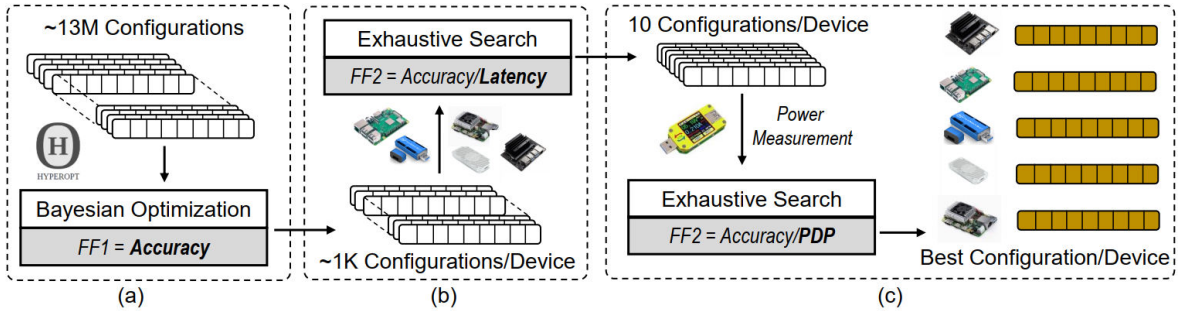

Figure 1: The proposed methodology to develop and optimize CNN models for facial expression recognition at the edge devices. (a) Bayesian optimization is used to shrink the search space from over 10M configurations to the top 1,000 configurations in terms of accuracy. (b) We use an exhaustive search on various edge devices to narrow down the search space to the top 10 CNN models in terms of accuracy/latency. (c) An exhaustive search is used to find the best CNN model for each edge AI accelerator in terms of accuracy/PDP.

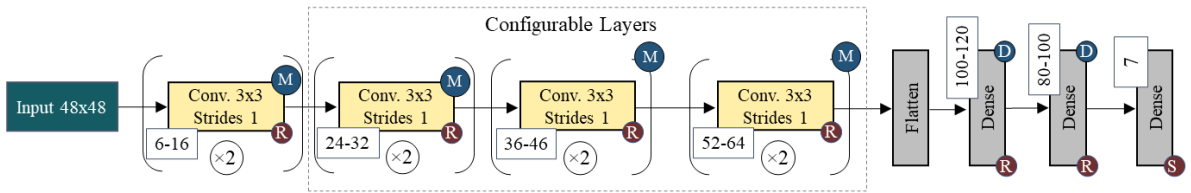

Figure 2: The network configuration. **R** ReLU, **M** MaxPooling2D, **D** Dropout, **S** Softmax.

Table 1: Network configuration settings

| Parameters | Description | Range | Step |
|---|---|---|---|
| Block | Number of blocks in the network | 2-4 | 1 |
| K1 | Number of first block's kernels | 6-16 | 2 |
| K2 | Number of second block's kernels | 24-32 | 4 |
| K3 | Number of third block's kernels | 36-48 | 4 |
| K4 | Number of fourth block's kernels | 52-64 | 4 |
| FC1 | Number of units in the first FC layer | 100-120 | 5 |
| DO1 | Probability of first dropout layer | 0.1-0.3 | 0.01 |
| FC2 | Number of units in the second FC layer | 80-100 | 5 |
| DO2 | Probability of second dropout layer | 0.1-0.3 | 0.01 |

## 2 PROPOSED METHODOLOGY TO DEVELOP AND OPTIMIZE CNN MODELS

In this paper, we propose a hierarchical approach for network configuration search, as shown in Fig. 1, which optimizes the CNN architectures for deployment on edge AI accelerators to achieve a balance between accuracy, latency, and power dissipation. Due to the different costs of measurements, we begin the configuration search with less expensive measurements and shrink the size of the search space at each step using different fitness functions.

The first stage of the proposed method for network optimization is shown in Fig. 1 (a), in which we use Hyperopt [12] to optimize the CNN architectures based on a single objective, i.e., accuracy. Hyperopt [12] is a Python library that offers automated hyperparameter tuning using various sequential model-based optimization (SMBO), a.k.a. Bayesian optimization, techniques. It supports various search algorithms including random search, simulated annealing, and Tree of-Parzen-Estimators (TPE), in which the search process is repeated for a number of iterations set by the user, and the best model is selected based on the loss function.

The general structure of the CNN models developed in this paper is inspired by the VGG architecture [16], as shown in Figure 2. Each model includes multiple VGG blocks containing two convolution layers with 3×3 kernel sizes and a stride of 1. The configurable parameters that are available for Hyperopt for optimizing the CNN architecture are listed in Table 1. In particular, (1) the number of VGG blocks (Block) can range between 2 and 4, allowing for varying network depths; (2) the number of kernels in each VGG block can change; (3) although the number of FC layers is kept constant, the number of nodes in each layer (except for the output layer) is variable, and finally (4) the dropout probability can vary between 0.1 and 0.3 with 0.01 intervals. The aforementioned configurable parameters create a search space with over 13 million different network configurations. Here, we prompt the Hyperopt to return over one thousand configurations in the search space that are optimized based on the accuracy fitness function. In this step, the cost of measurement, i.e., evaluating model accuracy, is in order of a few milliseconds per model.

In the second stage, shown in Fig. 1(b), we incorporate edge AI accelerators in the optimization approach and use Accuracy/Latency as the fitness function (FF). The new fitness function provides a balance between accuracy and latency. Here, we exhaustively search the configuration space, i.e., 1,000 models per device, and select 10 models with the highest fitness per device. In order to measure the latency, we run each model 40 times on each device independently and average the latency results. Depending on the edge device, it may take approximately 10 seconds to one minute to obtain the average latency.

The power consumption of each device is incorporated into the final stage of the proposed approach, as shown in Fig. 1(c).

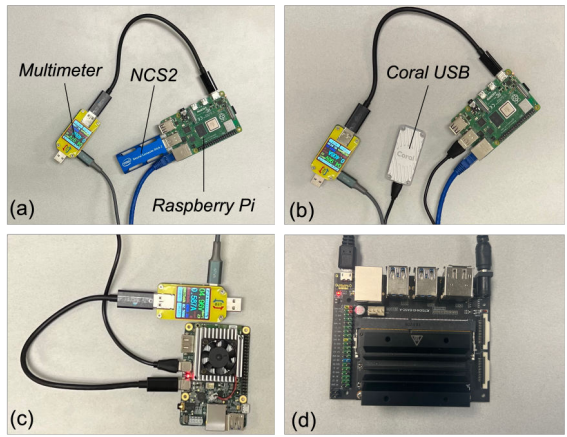

**Figure 3: Experimental setup. (a) Pi + NCS2 (b) Pi + Coral TPU (c) Coral Dev board (d) Jetson Nano.**

An exhaustive search is carried out on the given search space, which consists of 10 different configurations per device derived from the previous stage. The fitness function utilized in this stage is *Accuracy/PDP*, where PDP is the power-delay product. The power measurement is a manual process for all devices, except for Jetson Nano, as described in the next section. The time required to load and run the models to measure the average power consumption is approximately three to four minutes. At the end of this stage, the best configuration which provides a balance between accuracy, latency, and power is selected for each edge device.

## 3 EXPERIMENTAL SETUP

### 3.1 Dataset

The dataset we chose for this work is the *Extended Cohn-Kanade* (CK+) [17] consisting of 593 video sequences, each of which contains frames that capture subjects as their facial expressions shift from a neutral expression to one of seven facial expressions labeled as *happiness*, *sadness*, *surprise*, *anger*, *contempt*, *disgust*, or *fear*. The CK+ dataset was created from 123 different subjects originating from a wide range of different ages, genders, and heritages. This is a relatively small dataset, and there is some extra effort required up-front in order to extract the appropriate frames from the video sequences as well as authorization required by the creators of the dataset before it can be used in research. To construct an image dataset from the video sequences, we extract the peak frame from each video sequence. A *peak frame* is a frame that captures the subject's target facial expression. The peak frames for each video file are labeled by the creators of the dataset. Additionally, we applied a pre-processing step to resize the images to 48 × 48 pixels and convert them all to grayscale prior to training.

### 3.2 Edge AI Accelerators

In this work, we test our methodology to develop and optimize CNN models for facial expression recognition on various edge devices including Raspberry Pi, Google Coral TPU (both Dev board and USB), Intel Movidius neural compute stick 2 (NCS2), and Nvidia Jetson Nano, as shown in Fig. 3. Each edge device has its own characteristics and deployment procedures as described in the following.

*3.2.1 Nvidia Jetson Nano.* The NVIDIA Jetson Nano is a low-cost development board for ML applications. The Jetson Nano is a modular computer with a Tegra X1 SoC that combines an ARM A57 quad-core, a 1.43 GHz CPU, and four distinct 32-CUDA core processing blocks (128 CUDA cores total) within a Maxwell architecture. It also comes with 4 GB of RAM. Nvidia provides two operating modes for Jetson Nano, i.e., 5W and Max-N, that can be configured through a software interface. In the 5W mode, also known as low power mode, only two cores of the ARM A57 are powered on, the clock frequency is restricted to 0.9 GHz and the GPU's clock frequency is limited to 0.64 GHz. However, in the Max-N mode, the Arm 57's four active cores operate at a clock frequency of 1.5 GHz, while the GPU's clock frequency is 0.92 GHz. TensorRT models with 16-bit floating point (FP16) precision are used for Jetson Nano. We export the models from TensorFlow to ONNX and then import them into TensorRT using OnnxParser.

*3.2.2 Intel Neural Compute Stick 2 (NCS2).* The NCS2 is based on the Intel Movidius X Vision Processing Unit (VPU), which contains 16 programmable SHAVE cores and a dedicated neural compute engine for DNN inference acceleration. It is equipped with a 700 MHz base frequency and a 16 nm technology node. NCS2 also has 4 GB of RAM and a maximum frequency of 1600 MHz. NCS2 supports 16-bit floating point operations. Intel provides a Python3 and C library called OpenVINO for deploying ML models on the NCS2. OpenVINO comes with a model optimizer that converts the models to a format that allows for them to be deployed on NCS2.

*3.2.3 Google Coral TPU.* The edge TPU is used as a co-processor on Coral's Dev Board (TPU-DEV) together with the NXP i.MX 8M SoC on their system-on-module (SoM) architecture. In order to run on the Coral Edge TPU, all models must be converted to the TensorFlow Lite format and quantized to 8-bit integer types. Coral also provides a USB accelerator (TPU-USB) that can be integrated with a CPU as a co-processor. With a power usage of approximately two watts, this accelerator is ideal for low-power settings.

### 3.3 Power Measurement Setup

As shown in Fig. 3, to measure the power dissipation of all edge devices except for the Jetson Nano, we employed a USB 3.0 multimeter, i.e., MakerHawk UM34C. Due to its inline nature, this measurement device records the whole system's power utilization, including USB I/O. After connecting the power measuring instrument to an input power port, both the idle and running powers are recorded. First, idle power is measured per second for three minutes after connecting the edge devices to the multi-meter without running any models on the devices. Next, we run all models for 3 minutes on each device and measure the dynamic power by subtracting the recorded power from the idle state from the total power dissipated while running the models. To measure the power dissipation of Jetson Nano, we use its three internal sensors, which are located at the board's power input, the GPU, and the CPU. We read the sensors automatically using the *tegrastats* utility. For Jetson Nano, we run each model for three minutes and log the CPU and GPU power consumption.

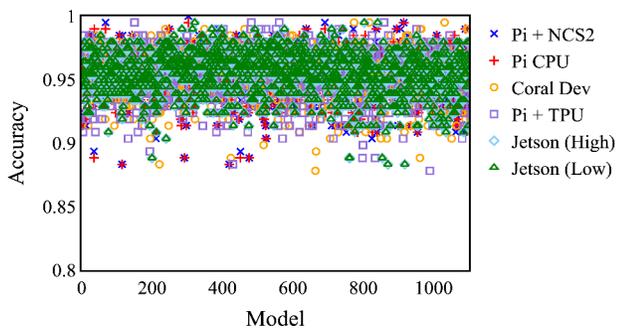

Figure 4: Accuracy for all CNN models deployed on each edge AI accelerator.

Table 2: Average and standard deviation for accuracy, latency, and dynamic power metrics for various devices.

| Device | Accuracy (%) | | Latency (ms) | | Power (W) | |
|---|---|---|---|---|---|---|
| | Ave | Std | Ave | Std | Ave | Std |
| Pi | **98.88** | 0.35 | 4.70 | 0.799 | 1.41 | 0.014 |
| Jetson Nano - L | 95.45 | 1.69 | 1.92 | 0.173 | 1.03 | 0.076 |
| Jetson Nano - H | 95.45 | 1.69 | 1.93 | 0.180 | 2.37 | 0.522 |
| Pi + NCS2 | 95.45 | 1.68 | 2.51 | 0.063 | 2.15 | 0.073 |
| Pi + TPU | **98.88** | 0.38 | 1.87 | 0.130 | 0.82 | 0.018 |
| Dev board | **98.88** | 0.38 | **0.47** | 0.043 | **0.55** | 0.013 |

## 4 RESULTS AND DISCUSSION

In this section, we provide comprehensive analyses and comparisons across various edge devices in terms of accuracy, latency, and power consumption as described in the following.

### 4.1 Accuracy

Figure 4 illustrates the range of accuracy, from 88.32% to 99.49%, achieved by over 1,000 CNN models chosen at the end of the first stage of the proposed optimization methodology. Each model is deployed on all the edge AI accelerators discussed in the previous section, leading to a total of over 5,000 experiments. As is shown, different accuracy levels, even for the same model, are offered since each device uses a different precision, such as Float32, Float16, and Int8. We further analyze all models running on each device by computing their average and standard deviation of the accuracy metric. As listed in Table 2, the highest average accuracy is 98.88, which is provided by three devices: the Raspberry Pi, the Raspberry Pi + Coral TPU, and the Coral Dev board. In terms of standard deviation, the mentioned edge devices are slightly different. It is worth mentioning that both Coral TPUs use only Int8 operations and still offer higher accuracy compared to Jetson Nano and NCS2, which support FP-16 operations. At this stage, the model with the highest accuracy is identified as the best model since it is the only factor examined by the fitness function. Table 3 demonstrates that, across all edge devices, there are two distinct configurations that provide the highest accuracy, equaling 99.49%. The first design has four VGG blocks, which add up to 11 layers. The second design has three VGG blocks, which add up to 9 layers. Details on these structures can be found in the Table 3.

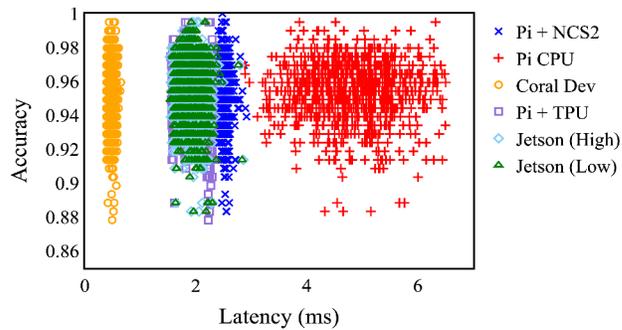

Figure 5: Accuracy vs. Latency for all CNN models deployed on each edge AI accelerator.

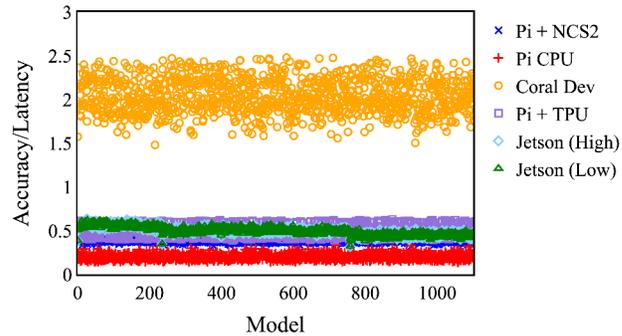

Figure 6: Accuracy/Latency ratio for all CNN models deployed on each edge AI accelerator.

### 4.2 Latency

The accuracy-to-latency ratio is the fitness function used in the second stage of our methodology. Various models may reach comparable accuracy levels, but the inference times may vary as a result of the models' configuration differences. Figure 5 shows the accuracy vs. latency for all models running on different devices. At first glance, it is clear that models running on the Raspberry Pi have the worst latency, while models running on the Coral Dev board have the lowest. Other devices with roughly comparable latency lie between these two. Figure 6 also shows the accuracy-to-latency ratio for all models running on edge devices. As shown, there is a significant difference between the results of the Coral Dev board and those of other devices, indicating that the Coral Dev board provides high-accuracy models while providing low latency. The mean and standard deviation of inference latency across all models deployed on each device are included in Table 2.

Table 3 provides the best configuration for each device based on accuracy/latency score. With the exception of the Raspberry Pi+TPU, models with just two VGG blocks, 7 layers total, have the highest score across all devices. Therefore, network depth is a key factor influencing the latency of the models. The order of the inference time from the lowest to the highest is as follows: Coral Dev board, Jetson Nano high power, Jetson Nano low power, Raspberry Pi+TPU, Raspberry Pi+NCS2, and Raspberry Pi. Separating the USB accelerators from development boards allows for better comparison as below.

*4.2.1 USB Accelerators.* According to Table 2, both the NCS2 and Coral USB accelerators achieve 1.87× and 2.51× average latency reduction compared to the baseline Raspberry Pi, respectively. Furthermore, as shown in Table 3, the best model in terms of accuracy/latency for the Coral USB accelerator has 26% lower latency and 1.52% higher accuracy than the best model on NCS2.

Table 3: Comprehensive Analysis of Different Devices

| Metric | Device | Best Model Architecture | | | | | | | | | Accuracy (%) | Latency (ms) | Power (W) |
|---|---|---|---|---|---|---|---|---|---|---|---|---|---|
| | | K1 | K2 | K3 | K4 | FC1 | DO1 | FC2 | DO2 | Output | | | |
| Accuracy | All Devices | 10 | 32 | 44 | 56 | 115 | 0.1 | 100 | 0.17 | 7 | 99.49 | - | - |
| | | 12 | 22 | 48 | - | 100 | 0.12 | 85 | 0.15 | 7 | 99.49 | - | - |
| Accuracy/Delay | Pi | 16 | 24 | - | - | 100 | 0.2 | 80 | 0.14 | 7 | 96.95 | 2.88 | - |
| | Jetson Nano - L | 10 | 32 | - | - | 120 | 0.29 | 80 | 0.19 | 7 | 97.46 | 1.57 | - |
| | Jetson Nano - H | 10 | 32 | - | - | 120 | 0.29 | 80 | 0.19 | 7 | 97.46 | 1.53 | - |
| | Pi + NCS2 | 16 | 24 | - | - | 100 | 0.2 | 80 | 0.14 | 7 | 97.46 | 2.35 | - |
| | Pi + TPU | 12 | 24 | 36 | - | 100 | 0.12 | 100 | 0.1 | 7 | **98.98** | 1.73 | - |
| | Coral Dev | 16 | 32 | - | - | 115 | 0.21 | 85 | 0.17 | 7 | 97.46 | **0.39** | - |
| Accuracy/PDP | Pi | 16 | 24 | - | - | 100 | 0.2 | 80 | 0.14 | 7 | 96.95 | 2.88 | 1.56 |
| | Jetson Nano - L | 10 | 28 | - | - | 120 | 0.11 | 85 | 0.18 | 7 | 97.46 | 1.62 | 0.91 |
| | Jetson Nano - H | 16 | 24 | - | - | 100 | 0.2 | 80 | 0.14 | 7 | 95.95 | 1.60 | 1.34 |
| | Pi + NCS2 | 16 | 24 | - | - | 100 | 0.2 | 80 | 0.14 | 7 | 97.46 | 2.35 | 2.08 |
| | Pi + TPU | 16 | 32 | - | - | 115 | 0.21 | 85 | 0.17 | 7 | 97.46 | 1.55 | 0.77 |
| | Coral Dev | 18 | 24 | - | - | 110 | 0.15 | 95 | 0.29 | 7 | 96.95 | **0.39** | **0.52** |

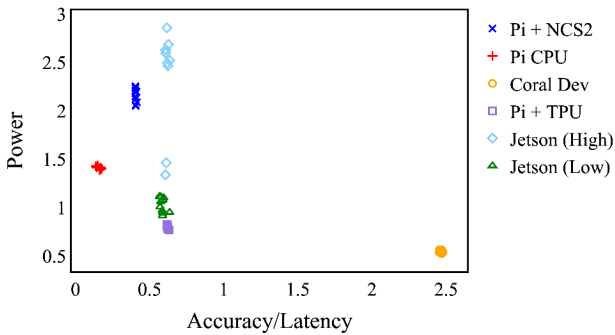

Figure 7: Power consumption for Top-10 models ranked by Accuracy/Latency metric.

*4.2.2 Development Boards.* Table 2 shows that both Coral TPU and Jetson Nano development boards outperform the Raspberry PI in terms of inference speed. Using the low and high power modes, Jetson Nano can achieve an average speedup of 2.43 and 2.44 over the Raspberry Pi, respectively. Also, a 10× latency reduction is achieved for the Coral Dev board compared to the baseline Raspberry Pi. Furthermore, as shown in table 3, the best model in terms of accuracy/latency for the Coral Dev board realizes 4.02× and 3.92× speedup compared to the Jetson Nano with low-power and high-power modes, respectively.

## 4.3 Power Consumption

The last metric we used in our methodology to evaluate the performance of the models was *Accuracy/PDP*, which is a combined measure of the accuracy, latency, and power consumption of each model. The results for the 10 best models are shown in Figure 7. As we can see, the Coral Dev board not only has the best accuracy-to-latency ratio but also dissipates less power compared to other devices. As a result, the Coral Dev board outperforms other edge accelerators in terms of accuracy/PDP. Other devices except Raspberry Pi are almost in the same range of accuracy-to-latency ratio. However, in terms of dynamic power, Raspberry Pi + TPU consumes less followed by Jetson nano low-power, Raspberry Pi + NCS2, and Jetson Nano High-power. Additionally, Table 2 lists the average dynamic power for the top ten models chosen in the final optimization phase. Finally, Table 3 lists the top model for each device based on *Accuracy/PDP* fitness function. The highest accuracy is achieved by Jetson Nano, Raspberry Pi + NCS2, and Raspberry Pi + TPU. Coral Dev board, however, outperforms other devices in terms of latency and dynamic power. We compare the USB accelerators and Dev boards separately for further analysis.

*4.3.1 USB Accelerators.* Table 2 shows that models running on the Raspberry Pi with the TPU use 2.62× less dynamic power than those running on the Raspberry Pi with the NCS2. Table 3 additionally shows that the best model in terms of accuracy/PDP deployed on Raspberry Pi + TPU achieves a 2.70× dynamic power reduction when compared to Raspberry Pi + NCS2.

*4.3.2 Development Boards.* Table 2 shows that models running on the Coral Dev board consume 4.30× and 1.87× less average power compared to Jetson Nano in high and low power modes, respectively. Additionally, as indicated in Table 3, the best model in terms of accuracy/PDP deployed on the Coral Dev board achieves power reduction of 2.58× and 1.75× compared to Jetson High and Low power, respectively.

## 4.4 Comparison with Previous Work

Here, we implemented two prior works to examine how our models perform compared to the previous work in the field. We only use the Coral Dev board since it provides the best overall results as verified in the previous subsections. We first implemented a similar VGG-like architecture proposed in [18]. Our top model on the Coral Dev board achieved an accuracy of 96.95%, which is only 0.51% lower than the accuracy attained by the CNN model proposed in [18]. It is worth noting that, the accuracy we achieved by deploying their model on the Coral Dev board is higher than the 96.62% accuracy reported in their paper. Our model uses nearly the same amount of dynamic power as their models while being 17.82× faster. The Accuracy/PDP metric was also compared to ensure a thorough analysis of the models. Table 4 demonstrates our model achieves roughly 17× improvement compared to theirs in terms of Accuracy/PDP. Additionally, we implemented an Inception-like model as described by Mollahosseini et. al.[19]. For a fair comparison, in our implementation, we removed the three color channels and only used gray-scale images for training and evaluation. When deploying their model on the Coral Dev board, we realized an accuracy of 95.93% which is again higher than the 93.2% reported in their paper [19]. Even with this increased accuracy, our best CNN model still achieves 1.53% higher accuracy. Furthermore, when compared to their approach, our model is more efficient in all metrics, including latency, power, and Accuracy/PDP, by a factor of 1.67, 1.29, and 2.17, respectively.

Table 4: Comparison with the previous FER models deployed on Coral TPU development board.

| Model | Accuracy (%) | Latency (ms) | Power (W) | Accuracy/PDP |
|---|---|---|---|---|
| [18] | 97.46 | 6.95 | 0.50 | 28.04 |
| [19] | 95.93 | 0.65 | 0.67 | 220.27 |
| Our Model | 96.95 | 0.39 | 0.52 | 478.06 |

## 5 CONCLUSION

In this work, we presented a methodology for the efficient deployment of CNN models on edge AI accelerators for facial expression recognition (FER). We use a hierarchical search method based on accuracy, latency, and power consumption to find the best model architecture. Due to the high cost of training models on edge devices, we performed hyperparameter optimization prior to deploying the models on edge devices by limiting the hyperparameter search space and using a VGG-inspired model architecture. We reduced the search space to over 1,000 models by first ranking the models based on the accuracy, then latency is used to further down-select before measuring power consumption on the highest performing model configurations across each of the edge devices used in this work. We reached a peak accuracy of 99.49% on the CK+ facial expression dataset. The model configurations found using the proposed optimization strategy show high rates of facial expression classification while consuming significantly lower power and energy than large-scale CNN models. Small-scale FER models provide benefits in many applications, including social robotics, where facial expression recognition tasks must be carried out in real-time on resource- and energy-constrained edge devices.